\documentclass{bmvc2k}
\usepackage{arydshln}
\graphicspath{{images/}}
\usepackage[outdir=./images/]{epstopdf}\usepackage{stackengine}
\usepackage{stackengine}
\newcommand{\red}[1]{\textcolor{black}{#1}}
\title{Domain Adaptation for Object Detection via Style Consistency}

 \addauthor{Adrian Lopez Rodriguez}{adrian.lopez-rodriguez15@imperial.ac.uk}{1}
 \addauthor{Krystian Mikolajczyk}{k.mikolajczyk@imperial.ac.uk}{2}

 \addinstitution{
  MatchLab\\
  Imperial College London\\
  London, UK
 }

\runninghead{Lopez, Mikolajczyk}{Domain Adaptation for Object Detection via Style}

\def\eg{\emph{e.g}\bmvaOneDot}
\def\ie{\emph{i.e}\bmvaOneDot}

\begin{document}

\maketitle

\begin{abstract}
We propose a domain adaptation approach for object detection. We introduce a two-step method: the first step makes the detector robust to low-level differences and the second step adapts the classifiers to changes in the high-level features. For the first step, we use a style transfer method for pixel-adaptation of source images to the target domain. We find that enforcing low distance in the high-level features of the object detector between the style transferred images and the source images improves the performance in the target domain. For the second step, we propose a robust pseudo labelling approach to reduce the noise in both positive and negative sampling. Experimental evaluation is performed using the detector SSD300 on PASCAL VOC extended with the dataset proposed in \cite{inoue2018cross}, where the target domain images are of different styles. Our approach significantly improves the state-of-the-art performance in this benchmark.
\end{abstract}


\maketitle

\section{Introduction}
\begin{figure}[t]
	\centering
	\includegraphics[trim={0.0cm 0.0cm 0cm 0cm},clip, width=\linewidth]{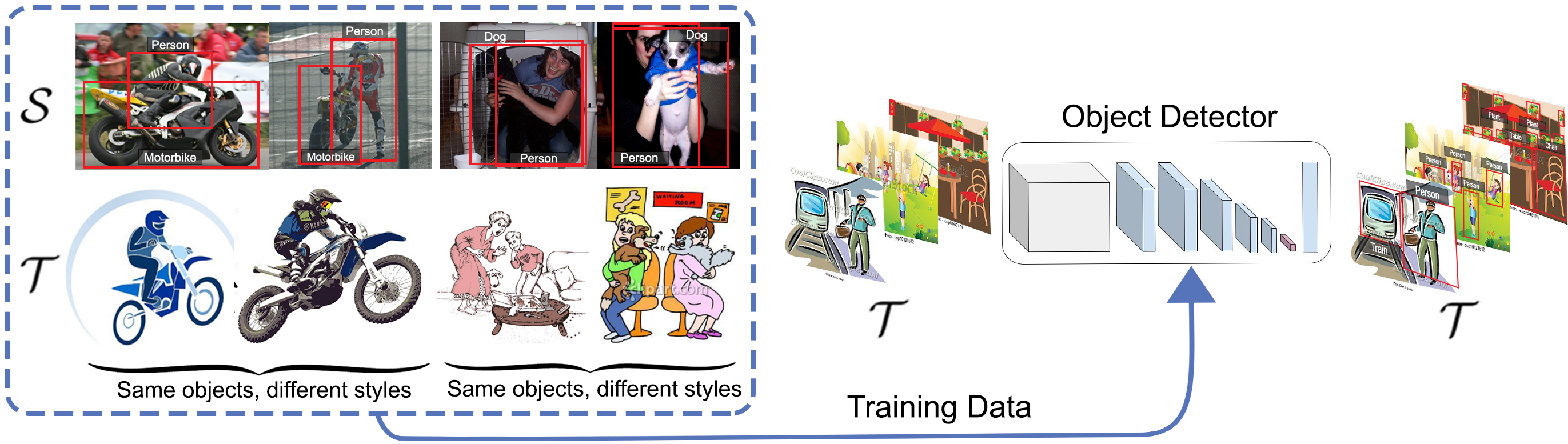}
	\caption{Overview of the unsupervised domain adaptation object detection task. We have a labelled source set $\mathcal{S}$ and an unlabelled target set $\mathcal{T}$ that shares classes with $\mathcal{S}$. We aim to use both sets to train a detector with good accuracy in $\mathcal{T}$.}
	\label{fig:problem}
	\vspace{-0cm}
\end{figure}
Object detection has been extensively studied in recent years, resulting in the development of highly accurate models such as Faster R-CNN \cite{ren2015faster} or SSD \cite{liu2016ssd}. Typically, object detectors rely on a generic backbone (\eg VGG \cite{DBLP:journals/corr/SimonyanZ14a} or ResNet~\cite{DBLP:journals/corr/HeZRS15}) pretrained in ImageNet \cite{deng2009imagenet} for classification tasks, and then finetuned on datasets with similar images for both training and evaluation, such as MS COCO \cite{lin2014microsoft} and PASCAL VOC \cite{everingham2010pascal}. However, such models do not behave well when trained and tested in different domains as shown in \cite{inoue2018cross, chen2018domain, saito2018strong}. \par
Several approaches have been developed for unsupervised domain adaptation, aiming to train a model using a fully-annotated source dataset $\mathcal{S}$ and a non-annotated target dataset $\mathcal{T}$ to perform well in $\mathcal{T}$. The proposed methods range from high-level feature adaptation via domain confusion  \cite{ganin2015unsupervised, tzeng2017adversarial} to image adaptation using a generative model~\cite{hoffman2017cycada, tzeng2018splat}.\par 
Image adaptation has been shown to be beneficial for closing the domain gap between the source and target domain, especially when the two domains are largely dissimilar in low-level statistics. The generative models usually employed for the domain adaptation task, such as CycleGAN \cite{zhu2017unpaired}, only provide one translated image per source image and cannot translate the source instance to a specific style in the target domain. Some datasets may contain several styles as shown in Figure \ref{fig:problem}, \eg changes in texture, shapes, patterns, illumination or colors. In those cases, having the capability of generating several translated source images to any of the target styles may help to close the domain gap. Style transfer methods are capable of adapting the input source image in an instance-level manner, making them suitable for the domain adaptation task. Apart from low-level changes, the object appearance can also vary at a high level, \eg shape changes. For good performance on the target domain, we also need to adapt the high-level features once the model is robust to the low-level changes, which can be done pseudo labelling the target domain instances using the detector trained in the source domain. \par
We propose an approach for domain adaptation of object detectors that relies on two main contributions. First, we use a style transfer method to adapt the source images to the target domain. We force feature consistency between the non-transferred image and the different styles, aiming to make the detector robust to low-level variations. Secondly, we propose robust pseudo labelling to provide labelled training examples from the unannotated target domain, using positive and negative example sampling with low label noise. The pseudo labels are used along with the style transfer method to train on style transferred target images. 

The remainder of the paper is organized as follows. We first review related works in the area of object detection, style transfer and domain adaptation in Section~\ref{sec:related_work}. Section \ref{sec:prelim} presents the style transfer method and the object detector used. We then introduce our approach in Section~\ref{sec:method}. Experimental results and discussions are in Section~\ref{sec:experiments}.
\section{Related work}\label{sec:related_work}
We present the relevant literature on general object detection task,  style transfer and domain adaptation.\par
\noindent\textbf{Object detectors} based on deep learning can be categorized in one-stage or two-stage methods. 

Two well-known two-stage detectors are R-CNN \cite{girshick2014rich} and Fast R-CNN \cite{Girshick_2015_ICCV}, both of which used a low-level vision algorithm for the first stage. Faster R-CNN \cite{ren2015faster} introduced a Region Proposal Network (RPN), making the model end-to-end trainable.

Examples of one-stage detectors are YOLO \cite{redmon2016you} and SSD~\cite{liu2016ssd}, which do not have a ROI proposal network, leading to a more efficient inference.\par 
Some works tackled the problem of weakly-supervised object detection \cite{cinbis2017weakly, bilen2016weakly}. A related problem was investigated in \cite{hoffman2014lsda, tang2016large, uijlings2018revisiting}, where  a set of classes with annotated bounding-boxes are used to detect another set of classes with only image-level labels available.\par

\noindent\textbf{Style Transfer} method was proposed in \cite{gatys2016image}, which minimized a content and a style loss. Further extensions, such as \cite{johnson2016perceptual, ulyanov2016texture, dumoulin2017learned} used a feed-forward method instead. Contrary to the previous approaches, in \cite{huang2017arbitrary} a method was proposed to generalize to unseen styles. Recent developments focused on adapting the style of a real image (\eg illumination or weather) to another real image \cite{li2018closed, yoo2019photorealistic}.\par

\noindent\textbf{Domain adaptation}  has been actively researched for image classification problems. The approaches developed in \cite{tzeng2015simultaneous, ganin2015unsupervised} sought to confuse a domain classifier to align the features. A generalized framework using an adversarial method and untied weight sharing was introduced in \cite{tzeng2017adversarial}. An adversarial loss was also used in \cite{saitomaximum} to force alignment of $\mathcal{T}$ and $\mathcal{S}$ samples. In \cite{shu2018dirt} Virtual Adversarial Training was used along with a teacher network to refine the decision boundaries. In \cite{saito2017asymmetric} a co-training approach was applied with three classifiers, trained with source and agreed pseudo labelled target examples. The consistency between predictions for transformations of the input data was used in \cite{roy2019unsupervised}. Joint pixel and feature adaptation was attempted in \cite{hoffman2017cycada} using a similar approach to CycleGAN \cite{zhu2017unpaired}.

Domain adaptation for object detection is more challenging and only recently gained some attention. Some works used synthetically generated data to train an object detector for real images \cite{hattori2015learning, peng2015learning}. A recent approach \cite{rame2018omnia} focused on merging different fully-annotated datasets in different domains and with different object classes. In \cite{chen2018domain}, instance-level and image-level adaptation was done for Faster R-CNN. In \cite{inoue2018cross} a two-step training method was proposed, where CycleGAN was applied for pixel-level adaptation of PASCAL VOC images to other domains, then used for training a detector on domain adapted  images. A new dataset for object detection in domain adaptation was also introduced in~\cite{inoue2018cross} with significantly different domain styles (comic, watercolor and clipart), sharing the object categories with PASCAL VOC. A recent approach in \cite{tzeng2018splat} developed a framework to adapt GANs for the pixel-level translation in an object detection setting. In \cite{saito2018strong}, the focal loss \cite{lin2018focal} was used to adapt the high-level features that were more similar between the source and target  domains. A style transfer approach that is related to our method was used in \cite{wu2018dcan} for semantic segmentation, however no style consistency on the feature level was enforced.\par

\section{Preliminaries}\label{sec:prelim}
In this section we present the style transfer method and the object detector used in our experiments. We use for most of our experiments PASCAL VOC \cite{everingham2010pascal} as $\mathcal{S}$ and the dataset proposed in \cite{inoue2018cross} as $\mathcal{T}$, which contains three domains: Clipart1k, Watercolor2k and Comic2k. In each of these domains, there is a large variation of individual styles in the different instances. In \cite{inoue2018cross}, image adaptation from VOC to each of these three domains was performed using a CycleGAN. We will refer to these adapted images as Domain Transferred Images (\textit{DTI}). A more detailed overview of the datasets is given in the supplementary material.\par
\noindent\textbf{Style Transfer via Adaptive Instance Normalization.}\label{sec:style_transfer} 
 We adopt the method from \cite{huang2017arbitrary} to transfer the style of an image $x_\mathcal{T}$ to the image $x_\mathcal{S}$ via a feed-forward network resulting in $x_{\mathcal{S}\rightarrow{\mathcal{T}}}$. A feature map is generated using a VGG19 pretrained in ImageNet up to the layer \texttt{relu4\_1} for target $f_v(x_\mathcal{T})$, and source image $f_v(x_\mathcal{S})$. Next, $f_v(x_\mathcal{S})$ undergoes an affine transformation to match the mean and variance of $f_v(x_\mathcal{T})$ resulting in $f'_v(x_\mathcal{S})$, which is passed to a decoder that generates $x_{\mathcal{S}\rightarrow{\mathcal{T}}}$. The style transfer loss is the sum of a content loss $\mathcal{L}_{Cont}$ and a style loss $\mathcal{L}_{Style}$:
 \begin{equation}
     \mathcal{L}_{\mathit{ST}} = \mathcal{L}_\mathit{Cont} + \mathcal{L}_\mathit{Style}
     \label{eq:stloss}
 \end{equation}
 The content loss is defined as
 $\mathcal{L}_\mathit{Cont}=\|f'_v(x_\mathcal{S})-f_v(x_{\mathcal{S}\rightarrow{\mathcal{T}}})\|^2$ where $f_v(x_{\mathcal{S}\rightarrow{\mathcal{T}}})$ is the VGG19 feature map for $x_{\mathcal{S}\rightarrow{\mathcal{T}}}$. The style loss $\mathcal{L}_\mathit{Style}$ is defined as the Euclidean distance between the means and the standard deviations of the feature maps of $x_\mathcal{T}$ and $x_{\mathcal{S}\rightarrow{\mathcal{T}}}$ in layers \texttt{relu1\_1}, \texttt{relu2\_1}, \texttt{relu3\_1} and \texttt{relu4\_1}.

\noindent\textbf{Single Shot Detector (SSD).}\label{sec:ssd}
We build our approach on SSD \cite{liu2016ssd} but other methods such as Faster R-CNN \cite{ren2015faster} or YOLO \cite{redmon2016you} can be used as our approach addresses the training methodology rather than a specific detector. SSD is a single stage approach with a set of anchor boxes from different feature maps. The loss used during training with image $x$ is:
\begin{equation}
   \mathcal{L}_{\mathit{SSD}}(x) = \frac{1}{N}\sum_{b}(\mathcal{L}_{conf}(b,s_c) + \alpha \mathcal{L}_{loc}(b,g))
\label{eq:SSDloss}
\end{equation}
where confidence loss $\mathcal{L}_{conf}$ is the cross-entropy loss of score $s_c$ for ground-truth class label $c$ and matched box $b$ (Equation~\ref{eq:confloss}) averaged over all $N$ positive matched boxes in image $x$. 
$\mathcal{L}_{loc}$ is a smooth $l_1$ loss between the predicted geometric parameters of box $b$ and the ground-truth box $g$ for positive examples. Hyperparameters in our experiments include $\alpha=1$ and ratio $\mbox{3:1}$ of negative to positive object boxes in a given image. More details can be found in \cite{liu2016ssd}. $\mathcal{L}_{conf}$ for a single box $b$ with predicted score $s_c$ for correct class label $c$ is defined as: 
\begin{equation}
\mathcal{L}_{conf}(b,s_c) = -\log({s_c})
\label{eq:confloss}
\end{equation}
where $s_c$ is the normalized score applying the softmax function. 

\begin{figure*}[t]
	\centering
	\includegraphics[width=
	1\linewidth]{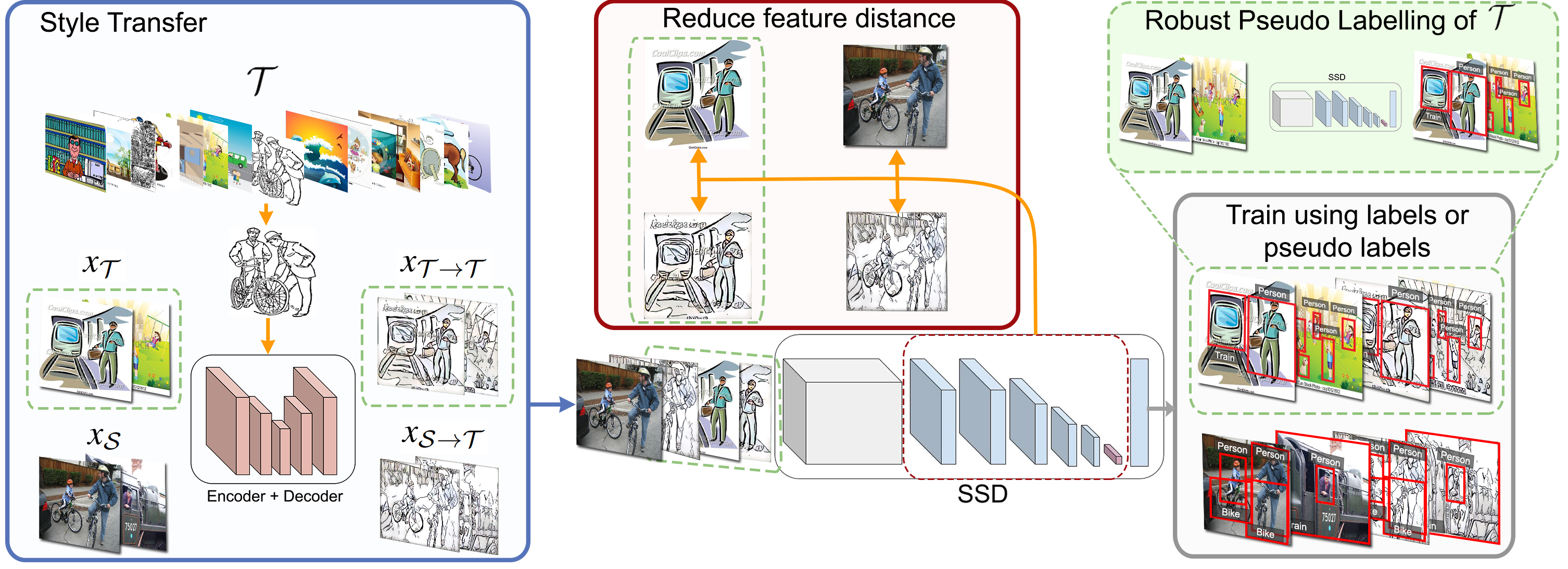}
	\vspace{-0.1cm}
	\caption{Overview of the proposed approach for domain adaptation of object detectors. We first make the detector robust to low-level variations by using a style transfer method for pixel-level adaptation of the source images, and then forcing feature consistency in the object detector between the style transferred image and the original image. Next, we pseudo label $\mathcal{T}$, and retrain following the same method with both images from $\mathcal{S}$ and $\mathcal{T}$.}
	\label{fig:method}
\end{figure*}

\section{Method}\label{sec:method}

In this section we present our approach, consisting of two steps. In the first step, we train the object detector via style transfer and feature consistency to improve robustness to low-level variations. To adapt it to high-level changes, we provide the detector with pseudo labelled training examples from $\mathcal{T}$ and finetune again. Figure \ref{fig:method} shows an overview of the method.
\subsection{Style Transfer with Feature Consistency}\label{sec:pixel_level_adapt}
Recent studies \cite{geirhos2018imagenettrained} have revealed that Deep Learning approaches are heavily reliant on low-level features for classification, making low-level adaptation crucial for domain adaptation. Thus, we use the style transfer approach presented in Section \ref{sec:style_transfer} for low-level adaptation. During training, we sample at random a style image $x_\mathcal{T}$ and use it to transfer each of the input images $x_{\mathcal{S}}$  into the target style, resulting in  $x_{\mathcal{S}\rightarrow{\mathcal{T}}}$ . Image $x_{\mathcal{S}\rightarrow{\mathcal{T}}}$ shares the ground-truth labels with $x_\mathcal{S}$, which we use to compute loss $\mathcal{L}_{\mathit{SSD}}$ for both $x_{\mathcal{S}\rightarrow{\mathcal{T}}}$ and $x_{\mathcal{S}}$. Low-level adaptation was done in \cite{inoue2018cross} with CycleGAN for obtaining one transferred image per $x_{\mathcal{S}}$, but with our method we are capable of generating several instances $x_{\mathcal{S}\rightarrow{\mathcal{T}}}$ out of the same $x_\mathcal{S}$.\par
\noindent\textbf{Enforcing feature consistency.} Our motivation to train the detector with $x_{\mathcal{S}\rightarrow{\mathcal{T}}}$  instances is to force the network to recognize the objects under various low-level transformations. However, no constraint is introduced on the feature level. Compared to $x_{\mathcal{S}}$, the generated $x_{\mathcal{S}\rightarrow{\mathcal{T}}}$ contain low-level variations, \eg color, texture or edges, but the location and shape of the objects is mostly preserved. Therefore, we assume that high-level (\ie semantic) features in the detector \red{should} remain invariant to the style change of the image. \red{Following that assumption, we introduce a constraint to reduce the distance between high-level features of $x_{\mathcal{S}\rightarrow{\mathcal{T}}}$ and $x_{\mathcal{S}}$ aiming to increase the robustness of high-level features to low-level changes.} Hence, the feature consistency loss is
\begin{equation}
    \mathcal{L}_{Cons} = \sum_{k=1}^K||f_k(x_{\mathcal{S}}) - f_k(x_{\mathcal{S}\rightarrow{\mathcal{T}}})||^2
    \label{eq:consloss}
\end{equation}
where $f_k(x)$ refers to one the last $K=6$ feature maps of SSD300, where the object detections are generated. Figure \ref{fig:method} shows how $\mathcal{L}_{Cons}$ is applied in the SSD300 detector.

\subsection{Robust Pseudo Labelling}\label{sec:pl}
The approach presented in Section \ref{sec:pixel_level_adapt} makes the high-level features extracted by the detector robust to low-level changes, \red{\ie the model extracts similar high-level features for both $x_\mathcal{S}$ and $x_\mathcal{S\rightarrow{\mathcal{T}}}$. However, the style transfer method does not incorporate any high-level semantic information in the image transfer process, meaning that there will still be a domain gap between the high-level features extracted by the detector from the images coming from $\mathcal{S}$ and the images coming from $\mathcal{T}$. We aim to adapt the classifiers to those high-level differences via a pseudo labelling approach.} A possible domain adaptation approach is to automatically label the target data using the source trained model $h_\mathcal{S}(x_{\mathcal{T}})$ and then obtain an improved  model $h_{\mathcal{S}+\mathcal{T}}$ by training with such pseudo labelled examples (refer to \cite{lee2013pseudo} for an in-depth explanation of why pseudo labels work). As past works \cite{cinbis2017weakly, zou2018unsupervised}, we use the confidence of $h_\mathcal{S}(x_{\mathcal{T}})$ to label $\mathcal{T}$. However, in an object detection problem, we also need to be careful when selecting negative examples.\par

\noindent{\textbf{Positive examples sampling}} has been done in \cite{inoue2018cross} by taking top-1 object per class in a target image in a weakly supervised setting, discarding other positive instances. We argue that the actual value of the label score $s_c=h(b)$ for box $b$ and class $c$ output by the detector can be used to filter out the less confident detections. We therefore propose to accept a pseudo labelled example if the probability of $c$ being correct is $P(c|h(b))\geq \epsilon_+$. The parameter $\epsilon_+$ defines the precision of the sampling process, which we expect to remain high in different domains for a given high $h(b)$, while the recall may drop significantly. We test this assumption by evaluating a detector trained on source VOC and tested in \textit{DTI} VOC domain, which present strong low-level feature changes. We report the results in the supplementary material. The precision curve is high for $\mathcal{S}$ and $DTI$ images \eg the detector confidence of $0.8$ corresponds to $P(c|(h(b)\geq 0.8))\geq 0.8$ in both domains, while recall drops from $0.6$ to $0.16$. Experimentally, we directly set the threshold $h(b)\geq \hat{s}_c$ that is then used to include the predictions as pseudo labelled examples. The pseudo labelling process is only run once.\par 
\noindent{\textbf{Negative examples sampling}} is equally important. 

Hard-negative mining during training of SSD in~\cite{liu2016ssd} is performed by choosing for a given instance the top-k boxes with the highest confidence $h(b_k)$ that were detected in the {background} regions, which gives good results when all ground-truth annotations are available. However, in the unsupervised setting with positive target examples pseudo labelled by $P(c|h(b))\geq \epsilon_+$ at high precision point, many false negative examples will have $P(c|h(b))$ just below $\epsilon_+$ therefore including the rejected top detections as negatives may result in positive examples within the negative set. Hence, we set another threshold $h(b)\geq \hat{s}_-$ for selecting the negative samples. We maintain the ratio of 3:1 negative to positive number of examples as in \cite{liu2016ssd} and negative sampling is performed for each instance during training.\par
\noindent{\textbf{Combining Pseudo Labels and Style Transfer.}} An image from the target domain, $x_{\mathcal{T}}$, can be transformed to have the style of another target image $x_{\mathcal{T}}$ using the style transfer model, generating a new instance $x_{\mathcal{T}\rightarrow{\mathcal{T}}}$, with the same but unknown labels as $x_{\mathcal{T}}$. The pseudo labels from $x_{\mathcal{T}}$ can be used with $x_{\mathcal{T}\rightarrow{\mathcal{T}}}$ for training the detector. Thus, the model detects high-level concepts in the target domain under different low-level variations, generated from individual image styles present in the target domain. \red{This approach is related to the consistency constraints applied in \cite{Roy_2019_CVPR}, however, instead of perturbations such as color jitter or adding noise, we apply a transformation in the target example to adapt it to the style of another target example.}
 Our loss for Robust Pseudo Labelling ${\mathcal{L}_{RPL}}$ has the same expression as  $\mathcal{L}_{SSD}$ loss (see Equation~\ref{eq:SSDloss}) but applied to  $x_{\mathcal{T}}$ and $x_{\mathcal{T}\rightarrow{\mathcal{T}}}$ instances where the pseudo labels are obtained with the strategy presented above.

\subsection{Combined Domain Adaptation Loss}\label{sec:combined}

Our final loss combines all the previously discussed terms:
\begin{equation}
\mathcal{L} = \lambda_1\mathcal{L}_{ST} + \lambda_2\mathcal{L}_{Cons} + \lambda_3\mathcal{L}_{\mathcal{S}} + \lambda_4\mathcal{L}_{\mathit{RPL}}
\label{eq:allloss}
\end{equation}

 In summary, our approach applies the style transfer loss $\mathcal{L}_{ST}$ (Equation \ref{eq:stloss}) enforcing consistency of VGG19 feature maps between $x_{\mathcal{S}}$ and $x_{\mathcal{S}\rightarrow{\mathcal{T}}}$ for the content loss. The consistency loss $\mathcal{L}_{Cons}$ (Equation \ref{eq:consloss})  performs similar operation on  high-level SSD feature maps  also between $x_{\mathcal{S}}$ and $x_{\mathcal{S}\rightarrow{\mathcal{T}}}$.  $\mathcal{L}_{\mathcal{S}}$ refers to detector loss $\mathcal{L}_{\mathit{SSD}}$ (Equation \ref{eq:SSDloss}) for both $x_{\mathcal{S}}$ and $x_{\mathcal{S}\rightarrow{\mathcal{T}}}$ images. Finally, $\mathcal{L}_{\mathit{RPL}}$ is also  $\mathcal{L}_{\mathit{SSD}}$ but for  $x_{\mathcal{T}}$ and $x_{\mathcal{T}\rightarrow{\mathcal{T}}}$ images based on our robust pseudo labelling approach (Section~\ref{sec:pl}). $\mathcal{L}_{ST}$ and $\mathcal{L}_{Cons}$ are also computed for $x_{\mathcal{T}}$ and $x_{\mathcal{T}\rightarrow{\mathcal{T}}}$ in the second step of training, \ie after pseudo labelling. $\lambda_1$, $\lambda_2$, $\lambda_3$ and $\lambda_4$ are hyperparameters.

\begin{table*}[ht]
	\scriptsize
	\setlength\tabcolsep{1pt}
	\centering
	\scalebox{0.96}{
	\begin{tabular}{*{22}{c}}    \toprule
		\multicolumn{22}{c}{\bfseries{Pascal VOC$\rightarrow{\text{Clipart1k}}$}}\\
		\toprule
		\emph{Model} & aero & bike &  bird & boat & bttle & bus &  car & cat & chair & cow  & table & dog & horse & mbike & prsn & plnt & sheep & sofa & train & tv & \textbf{mAP} \\\midrule
		\textit{Fully supervised}\\
		\text{$\mathcal{T}$ \cite{inoue2018cross}} & 50.5 & 60.3 & 40.1 & 55.9 & 34.8 & 79.7 & 61.9 & 13.5 & 56.2 & 76.1 & 57.7 & 36.8 & 63.5 & 92.3 & 76.2 & 49.8 & 40.2 & 28.1 & 60.3 & 74.4 & 55.4 \\
		\midrule
		\textit{Weakly supervised}\\
		\textit{{DTI}+{PL}} \cite{inoue2018cross} & 35.7 & 61.9 & 26.2 & 45.9 & 29.9 & 74.0 & 48.7 & 2.8 & 53.0 & 72.7 & 50.2 & 19.3 & 40.9 & 83.3 & 62.4 & 42.4 & 22.8 & 38.5 & 49.3 & 59.5 & 46.0 \\
		\midrule
		\textit{Unsupervised} & & & & &\\
		\text{$\mathcal{S}$ \cite{liu2016ssd}} & 19.8 & 49.5 & 20.1 & 23.0 & 11.3 & 38.6 & 34.2 & 2.5 & 39.1 & 21.6 & 27.3 & 10.8 & 32.5 & 54.1 & 45.3 & 31.2 & 19.0 & 19.5 & 19.1 & 17.9 & 26.8\\
		\textit{ADDA} \cite{tzeng2017adversarial} & 20.1 & 50.2 & 20.5 & 23.6 & 11.4 & 40.5 & 34.9 & 2.3 & 39.7 & 22.3 & 27.1 & 10.4 & 31.7 & 53.6 & 46.6 & 32.1 & 18.0 & 21.1 & 23.6 & 18.3 & 27.4 \\
    	\textit{DTI} \cite{inoue2018cross} & 23.3 & \textbf{60.1} & 24.9 & 41.5 & 26.4 & 53.0 & 44.0 & 4.1 & 45.3 & \text{51.5} & 39.5 & 11.6 & 40.4 & 62.2 & 61.1 & 37.1 & 20.9 & 39.6 & 38.4 & 36.0 & 38.0 \\
    	\textit{DA-Faster}* \cite{chen2018domain} & 15.0 & 34.6 & 12.4 & 11.9 & 19.8 & 21.1 & 23.2 & 3.1 & 22.1 & 26.3 & 10.6 & 10.0 & 19.6 & 39.4 & 34.6 & 29.3 & 1.0 & 17.1 & 19.7 & 24.8 & 19.8 \\
    	\textit{Strong-Weak}* \cite{saito2018strong} & 26.2 & 48.5 & \textbf{32.6} & 33.7 & \textbf{38.5} & 54.3 & 37.1 & \textbf{18.6} & 34.8 & \textbf{58.3} & 17.0 & 12.5 & 33.8 & 65.5 & 61.6 & \textbf{52.0} & 9.3 & 24.9 & \textbf{54.1} & 49.1 & 38.1 \\
        
        \hdashline
        \textbf{Ours}\\
        \textit{ST} & 32.0 & 56.9 & 24.9 & 36.0 & 29.3 & 51.2 & 46.7 & 4.0 & 47.8 & 55.7 & 38.9 & 16.1 & 43.7 & 79.2 & 64.2 & 39.2 & \textbf{21.8} & 43.3 & 38.6 & 47.9 & 40.9 \\
		\textit{ST+C} & 35.0 & 57.3 & 24.7 & 41.9 & 28.0 & 56.8 & 49.1 & 9.9 & 49.3 & 55.6 & 44.0 & \textbf{16.5} & 42.3 & 83.1 & 65.0 & 42.8 & 17.7 & 43.9 & 42.0 & 52.6 & 42.9\\
		\textit{ST+C+PL$^\dagger$ \cite{inoue2018cross}} & 34.8 & 51.9 & 23.2 & 43.8 & 29.8 & 58.6 & 48.2 & 6.9 & 46.1 & 56.2 & 40.6 & 20.0 & 37.6 & 84.2 & 60.3 & 40.2 & 19.9 & 37.8 & 34.6 & 48.8 & 41.2\\
		\textit{ST+C+PL \cite{inoue2018cross}} & 32.4 & 50.2 & 22.9 & 43.2 & 28.6 & 60.2 & 46.3 & 5.6 & 46.7 & 57.1 & 40.9 & 21.4 & 35.7 & 87.9 & 63.3 & 41.4 & 19.4 & 37.8 & 35.2 & 54.3 & 41.5\\
		\textit{ST+C+RPL$^\dagger$} & 35.5 & 58.8 & 24.9 & 41.9 & 25.3 & 59.1 & 48.4 & 10.2 & 49.4 & 55.7 & 41.5 & 15.9 & 42.4 & \textbf{84.3} & 66.1 & 42.2 & 18.7 & 43.8 & 44.0 & 52.6 & 43.0\\
		\textit{ST+C+RPL} & \textbf{36.9} & 55.1 & 26.4 & \textbf{42.7} & 23.6 & \textbf{64.4} & \textbf{52.1} & 10.1 & \textbf{50.9} & 57.2 & \textbf{48.2} & 16.2 & \textbf{45.9} & 83.7 & \textbf{69.5} & 41.5 & 21.6 & \textbf{46.1} & 48.3 & \textbf{55.7} & \textbf{44.8}\\

		\bottomrule
		\hline
	\end{tabular}}
	\vspace{1em}
	\caption{Object detection results in Average Precision (\%) in the domain Clipart1k. We report results for different combinations of the loss proposed in Section~\ref{sec:method}, and we use $\mathcal{S}$ for source domain data,  $\mathcal{T}$ for target, \textit{DTI} for domain transferred images, \textit{ST} for pixel level adaptation using style transfer, \textit{C} for consistency loss, \textit{PL} for the pseudo labelling method from \cite{inoue2018cross} and \textit{RPL} for robust pseudo labelling ($\dagger$~means no style transfer applied to $\mathcal{T}$ examples). Methods marked with * use a Faster R-CNN with images of 600 pixels on the shorter side. Our approach brings significant improvements over baseline and other state-of-the-art methods from \cite{inoue2018cross,tzeng2017adversarial,chen2018domain, saito2018strong}.}
	\label{tab:results_clipart}
\end{table*}
\begin{table}[ht]
	\footnotesize
	\setlength\tabcolsep{0.9pt}
	\begin{subtable}{0.54\textwidth}\centering
	\scalebox{0.71}{
	\begin{tabular}{*{8}{c}}    \toprule
		\ & \multicolumn{7}{c}{\bfseries{Pascal VOC$\rightarrow{\text{Watercolor2k}}$}}\\
		\toprule
		\emph{Model} & bike & bird & car &  cat & dog & prsn & \textbf{mAP} \\\midrule
		\textit{Fully sup.}\\
		$\mathcal{T}$\cite{inoue2018cross} & 76.0 & 60.0 & 52.7 & 41.0 & 43.8 & 77.3 & 58.4 \\ \midrule
		\textit{Weakly sup.}\\
		\textit{DTI+PL}\cite{inoue2018cross} & 76.5 & 54.9 & 46.0 & 37.4 & 38.5 & 72.3 & 54.3 \\
		\midrule
		\textit{Unsup.}\\
		$\mathcal{S}$ \cite{liu2016ssd} & 79.8 & 49.5 & 38.1 & 35.1 & 30.4 & 65.1 & 49.6\\
		\textit{DA-Faster}*\cite{chen2018domain} & 75.2 & 40.6 & 48.0 & 31.5 & 20.6 & 60.0 & 46.0 \\
		\textit{Str.-weak}*\cite{saito2018strong} & 82.3 & 55.9 & 46.5 & 32.7 & 35.5 & 66.7 & 53.3 \\
		\textit{DTI}\cite{inoue2018cross} & \textbf{82.8} & 47.0 & 40.2 & 34.6 & 35.3 & 62.5 & 50.4 \\

		\hdashline
		\textbf{Ours}\\
		\textit{ST} & 78.2 & 55.7 & 47.0 & 41.2 & 33.0 & 67.0 & 53.7\\
		\textit{ST+C} & 81.4 & 54.3 & 47.5 & 40.5 & 35.7 & 68.3 & 54.6\\
		\textit{ST+C+RPL}$^\dagger$ & 81.3 & 56.1 & 48.2 & 41.0 & 40.4 & 72.0 & 56.5\\
		\textit{ST+C+RPL} & 79.9 & \textbf{56.5} & \textbf{48.6} & \textbf{42.1} & \textbf{42.9} & \textbf{73.7} & \textbf{57.3}\\
		\bottomrule
		\hline
	\end{tabular} \hspace{-.3em}
	\begin{tabular}{*{7}{c}}    \toprule
		\multicolumn{7}{c}{\bfseries{Pascal VOC$\rightarrow{\text{Comic2k}}$}}\\
		\toprule
		bike & bird & car &  cat & dog & prsn & \textbf{mAP} \\\midrule
		\\
		55.9 & 26.8 & 40.4 & 42.3 & 43.0 & 70.1 & 46.4 \\\midrule
		\\
		55.2 & 18.5 & 38.2 & 22.9 & 34.1 & 54.5 & 37.2\\\midrule
		\\
	    43.9 & 10.0 & 19.4 & 12.9 & 20.3 & 42.6 & 24.9 \\
	    - & - & - & - & - & - & -\\
	    29.6 & \textbf{20.1} & 26.3 & 20.3 & 22.3 & 51.6 & 28.4\\
		43.6 & 13.6 & 30.2 & 16.0 & 26.9 & 48.3 & 29.8 \\

        \hdashline
		\\
		
		49.5 & 16.4 & 37.6 & 20.8 & 30.9 & 57.1 & 35.4\\ 
		51.4 & 17.3 & 39.9 & 21.4 & \textbf{31.9} & 56.1 & 36.3\\
		55.0 & 18.4 & 40.9 & 22.7 & 31.5 & 59.5 & 38.0\\
		\textbf{55.9} & 19.7 & \textbf{42.3} & \textbf{23.6} & 31.5 & \textbf{63.4} & \textbf{39.4}\\
		\bottomrule
		\hline
	\end{tabular}}
	\caption{Results in Watercolor2k and Comic2k}
	\label{tab:results_watcom}
	\end{subtable}
	\begin{subtable}{0.4\textwidth}\centering\vspace{0.62em}
    \quad\hspace{-0.2cm}
	\scalebox{0.71}{
	\begin{tabular}{*{2}{c}}    \toprule
		\multicolumn{2}{c}{\bfseries{Sim10k$\rightarrow{\text{Cityscapes}}$}}\\
		\toprule
		\emph{Model} & \multicolumn{1}{c}{\textbf{car}} \\
		\midrule
		\textit{Fully sup.}\\
		$\mathcal{T}$ & 60.1 \\ \midrule
		\textit{Unsup.}\\
        $\mathcal{S}$ & 43.7\\
        $\mathcal{S}$*\cite{saito2018strong} & 34.6\\
        \textit{DA-Faster}*\cite{chen2018domain} & 38.9\\
        \textit{Str.-weak}*\cite{saito2018strong} & 42.3\\
        \hdashline
		\textbf{Ours}\\
        \textit{ST} & 43.7 \\
        \textit{ST+C} & 44.0\\
        \textit{ST+RPL} & 44.1\\
        \textit{ST+C+RPL$^\dagger$} & 42.1 \\
        \textit{ST+C+RPL} &\textbf{44.2}\\
		\bottomrule
		\hline
		
	\end{tabular} \hspace{-0.55em}
	\scalebox{1}{
	
	\begin{tabular}{*{9}{c}}    \toprule
     \multicolumn{8}{c}{\bfseries{Cityscapes$\rightarrow{\text{Foggy Cityscapes}}$}}\\
		\toprule
		bike & bus & car & mbike & prsn & rider & train & truck  & \textbf{\ \ mAP} \\
        \midrule
        \\
        35.4 & 53.3 & 56.9 & 31.2 & 28.6 & 35.4 & 43.6 & 34.0 & 39.8 \\ \midrule \\
        29.1 & 28.5 & 38.8 & 23.8 & 22.1 & 25.7 & 18.8 & 17.6 & 25.6\\
        26.5 & 22.3 & 34.3 & 15.3 & 24.1 & 33.1 & 3.0 & 4.1 & 20.3\\
        31.0 & 25.0 & 40.5 & 22.1 & 35.3 & 20.2 & 20.0 & 27.1 & 27.6\\
        \textbf{35.3} & \textbf{36.2} & \textbf{43.5} & \textbf{30.0} & \textbf{29.9} & \textbf{42.3} & \textbf{32.6} & \textbf{24.5} & \textbf{34.3}\\
        \hdashline\\
        29.7 & 32.6 & 40.8 & 26.1 & 24.1 & 30.3 & 20.9 & 23.4 & 28.5\\
        30.1 & 32.2 & 39.6 & 25.6 & 24.1 & 29.8 & 23.0 & 23.0 & 28.4\\
        29.6 & 32.3 & 42.1 & 26.9 & 24.4 & 30.2 & 24.3 & 22.4 & 29.0\\
        28.4 & 31.3 & 39.0 & 23.4 & 22.0 & 24.7 & 21.1 & 22.0 & 26.5\\
        30.3 & 35.4 & 41.5 & 26.9 & 24.2 & 29.2 & 26.7 & 23.1 & 29.7\\
		\bottomrule
		\hline
	\end{tabular}}}\vspace{0.58em}
	\caption{Results in driving datasets}
	\label{tab:results_car}
	\end{subtable}
	\vspace{0.8em}
	\caption{Average Precision (\%) in Watercolor2k and Comic2k, and in driving datasets.}
	\vspace{-0.23cm}
\end{table}
\section{Experiments}\label{sec:experiments}\vspace{-0.02cm}
\noindent\textbf{Architecture.} We use the ChainerCV \cite{niitani2017chainercv} implementation of SSD300, which employs a VGG16 as feature extractor and is pretrained on the \textit{trainval} subset of VOC2007 and VOC2012. The input images are resized to $300$x$300$ and during inference we discard regions with a confidence lower than $0.01$. An intersection over union (IoU) threshold of $0.45$ is used for Non-Maximum Suppression as in \cite{liu2016ssd}. For the Style Transfer model we follow the  implementation of \cite{huang2017arbitrary}, where the encoder is a VGG19 pretrained on ImageNet and frozen during training. The decoder mirrors the encoder and is pretrained using WikiArt \cite{nichol2016wikiart} as style images and MS COCO \cite{lin2014microsoft} as content images.\par
\noindent\textbf{Training details.} We finetune the detector using SGD with a learning rate of $10^{-5}$ and a momentum of $0.9$ as in \cite{inoue2018cross}. The training is done in two steps. In the first step, we set $\lambda_1=1.0$, $\lambda_3=1.0$ and $\lambda_4=0$, and $\lambda_2=1.0$ or $\lambda_2=0.0$ depending if feature consistency is enforced. Before the second step, we pseudo label the target set with the updated object detector, and train again setting $\lambda_4=1.0$. We only pseudo label $\mathcal{T}$ once, using a threshold of $\hat{s}_c = 0.7$ (see Section~\ref{sec:pl}). For negative sampling, we set ${\hat{s}_{-} = 0.9}$. With the higher $\hat{s}_{-}$ compared to $\hat{s}_c$ we aim to avoid including many false negatives as, contrary to positive sampling, negative sampling is performed during training by selecting the hardest negative boxes satisfying $h(b)\geq \hat{s}_-$. However, the method is stable to lower values of $\hat{s}_{-}$ as discussed later in this section. In the first step, the batch only contains $\mathcal{S}$ instances, and in the second step half of the examples come from $\mathcal{S}$ and half from $\mathcal{T}$. \red{We use examples coming from $\mathcal{S}$ also in the second step to mitigate the effects of the incorrect pseudo-labels present in $\mathcal{T}$}. The batch size is $6$ and we finetune the detector for $10000$ iterations in the first step and for $5000$ iterations in the second step. Due to the small number of test images, and few instances for some classes, the obtained mean Average Precision (mAP) has a high variance. Thus, we perform mAP evaluation in the test set every $100$ iterations and we report the average mAP of the last $10$ evaluations.\par
\noindent\textbf{Comparison.} Our main experiments are performed in the cross-domain dataset  \cite{inoue2018cross} mentioned in Section \ref{sec:prelim}. We report the per-class average precision and mAP using the evaluation code in ChainerCV as \cite{inoue2018cross}. We include the results from \cite{inoue2018cross} for the weakly supervised method, referred to as \textit{DTI+PL}, and the ablation study of using only the \textit{DTI}. For Clipart1k and Watercolor2k, we include the results from \cite{saito2018strong} for their method, \textit{Strong-Weak}, and for \cite{chen2018domain}, \textit{DA-Faster}. Both \cite{chen2018domain} and \cite{saito2018strong} use Faster R-CNN with images of 600 pixels on the shorter side. For Comic2k we run the official code of \cite{saito2018strong} with global and local adaptation and context regularization. We also include the results for \textit{ADDA} \cite{tzeng2017adversarial} in Clipart1k reported in \cite{inoue2018cross}.\par

\subsection{Domain Adaptation for Object Detection Results}
Table \ref{tab:results_clipart} shows the results obtained in Clipart1k and Table \ref{tab:results_watcom} in Watercolor2k and Comic2k. Compared to \textit{DTI}, the performance increases in all three datasets by using the style transfer method, especially in Comic2k with an increase of 5.6\%. The improved performance shows that generating multiple $x_{\mathcal{S}\rightarrow{\mathcal{T}}}$ from a single $x_{\mathcal{S}}$ is beneficial in datasets with various styles. Applying the consistency loss proves to be favourable in the three datasets, notably in Clipart1k we achieve an increase of 2.0\%. We explored the impact of sampling different styles by restricting the style to a randomly chosen single image from $\mathcal{T}$ in Clipart1k (with consistency loss) achieving 36.3\%, that is 6.3\% lower compared to using all of the styles. In the three datasets, using $x_{\mathcal{T}\rightarrow{\mathcal{T}}}$ instances along with their pseudo labels leads to an improved performance compared to using only $x_{\mathcal{T}}$ instances, referred to as \textit{ST+C+RPL}$^\dagger$. \red{Even though the pseudo labelling method proposed in \cite{inoue2018cross} was developed for the case where image-level labels were available, we also tested its performance for the unsupervised case, \ie taking the top-1 in confidence for each of the classes and not using any threshold for the negative sampling. We show in Table \ref{tab:results_clipart} that using the pseudo labelling method from \cite{inoue2018cross} for our second step results in lower performance compared to not using any pseudo labels. The drop in performance is due to three reasons. First, by only pseudo labelling the top-1 prediction per class, a high number of predictions with high-confidence is discarded. Second, the method from \cite{inoue2018cross} does not impose any constraint in the negative sampling process, resulting in selecting a high amount of false negatives during training. Third, when no image-level labels are given, the method from \cite{inoue2018cross} assumes that every class is present in the image, which increases the number of false positives.}\par
\par
\begin{figure}[t]
	\centering
	\begin{subfigure}[b]{0.49\linewidth}\centering
	\includegraphics[trim={0.0cm 0.0cm 0cm 0cm},clip, width=0.88\linewidth]{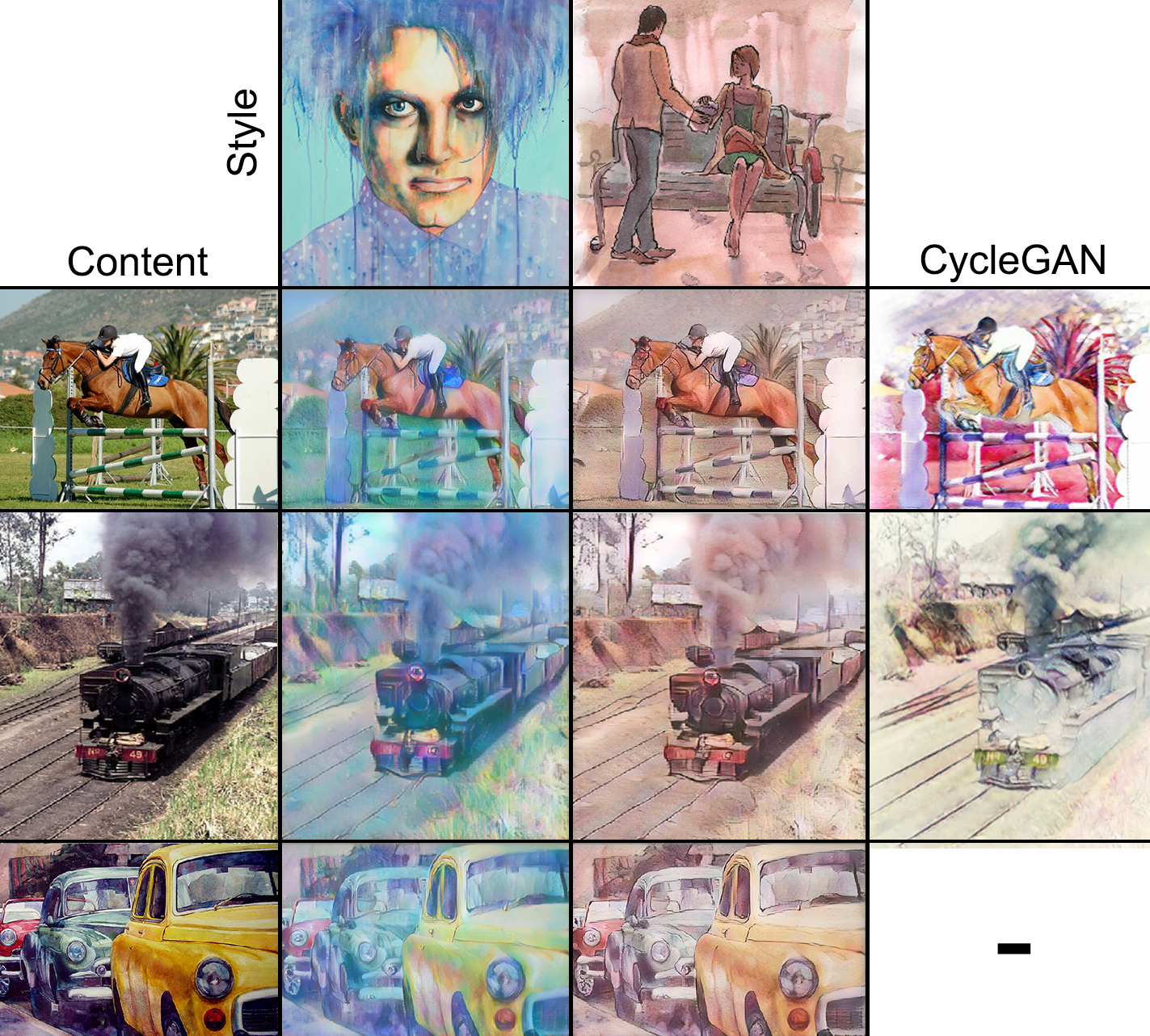}
	\caption{Examples of style transferred images}
	\end{subfigure}
	\begin{subfigure}[b]{0.49\linewidth}\centering
	\includegraphics[width=1\linewidth]{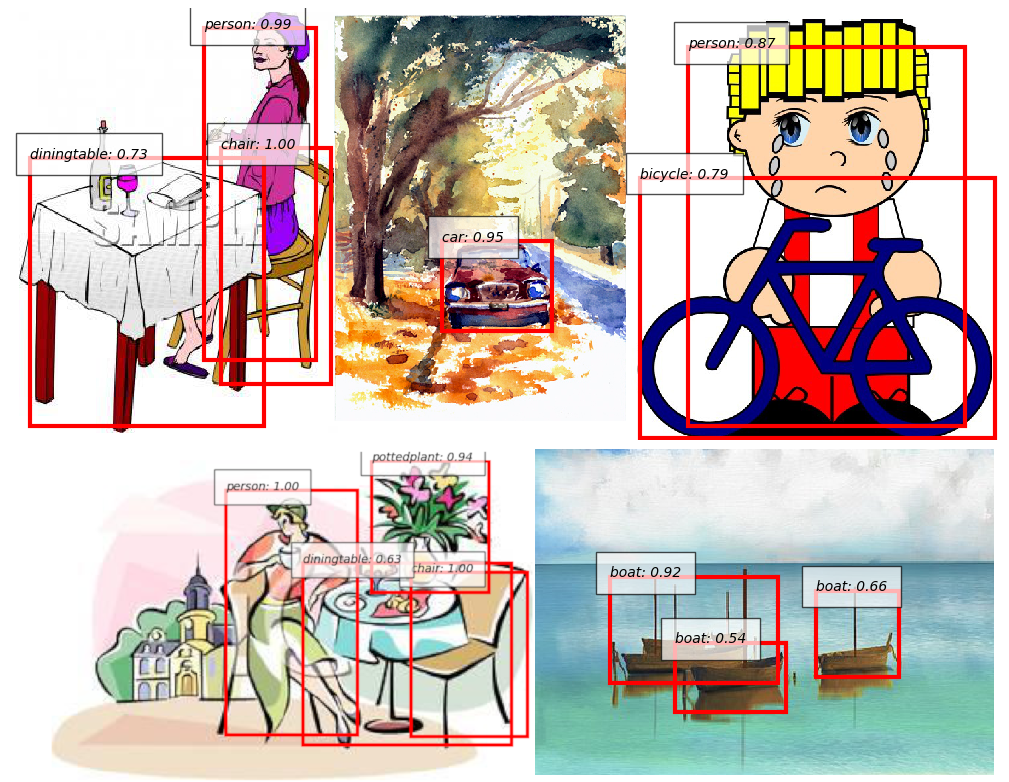}
	\caption{Examples of detections in the test set of $\mathcal{T}$}
	\end{subfigure}
	\caption{Qualitative results. Left: Examples of style transferred images along with their correspoding CycleGAN generated image in \cite{inoue2018cross}. Right: Examples of object detections in the target domain.}
	\label{fig:style_samples}
	\vspace{-0.1cm}
\end{figure}

Using all of our proposed modifications we obtain a model with a  higher mAP performance than the  state-of-the-art methods on the three datasets: by 6.7\% in Clipart1k, 4.0\% in Watercolor2k and 9.6\% in Comic2k. We also improve upon the weakly supervised method in \cite{inoue2018cross} for both Watercolor2k and Comic2k. In particular, in Watercolor2k the performance is only 1.1\% lower compared to the fully supervised approach.\par

\noindent{\bf Qualitative results.} In Figure \ref{fig:style_samples} we show some $x_{\mathcal{S}\rightarrow{\mathcal{T}}}$ samples, along with the corresponding image generated by CycleGAN. The last row shows an example of $x_{\mathcal{T}\rightarrow{\mathcal{T}}}$. Figure \ref{fig:style_samples} also shows examples of detections in $\mathcal{T}$. The model detects objects with  similar shapes but different low-level statistics than the $\mathcal{S}$ domain, \eg the car in the upper row, objects with shape changes, \eg the person in the bottom-left image, or objects with a simplified shape, \eg the bicycle in the upper-right image.\par
\begin{figure}[t]
	\begin{subfigure}[b]{0.49\linewidth}\centering
	\includegraphics[width=0.9\linewidth]{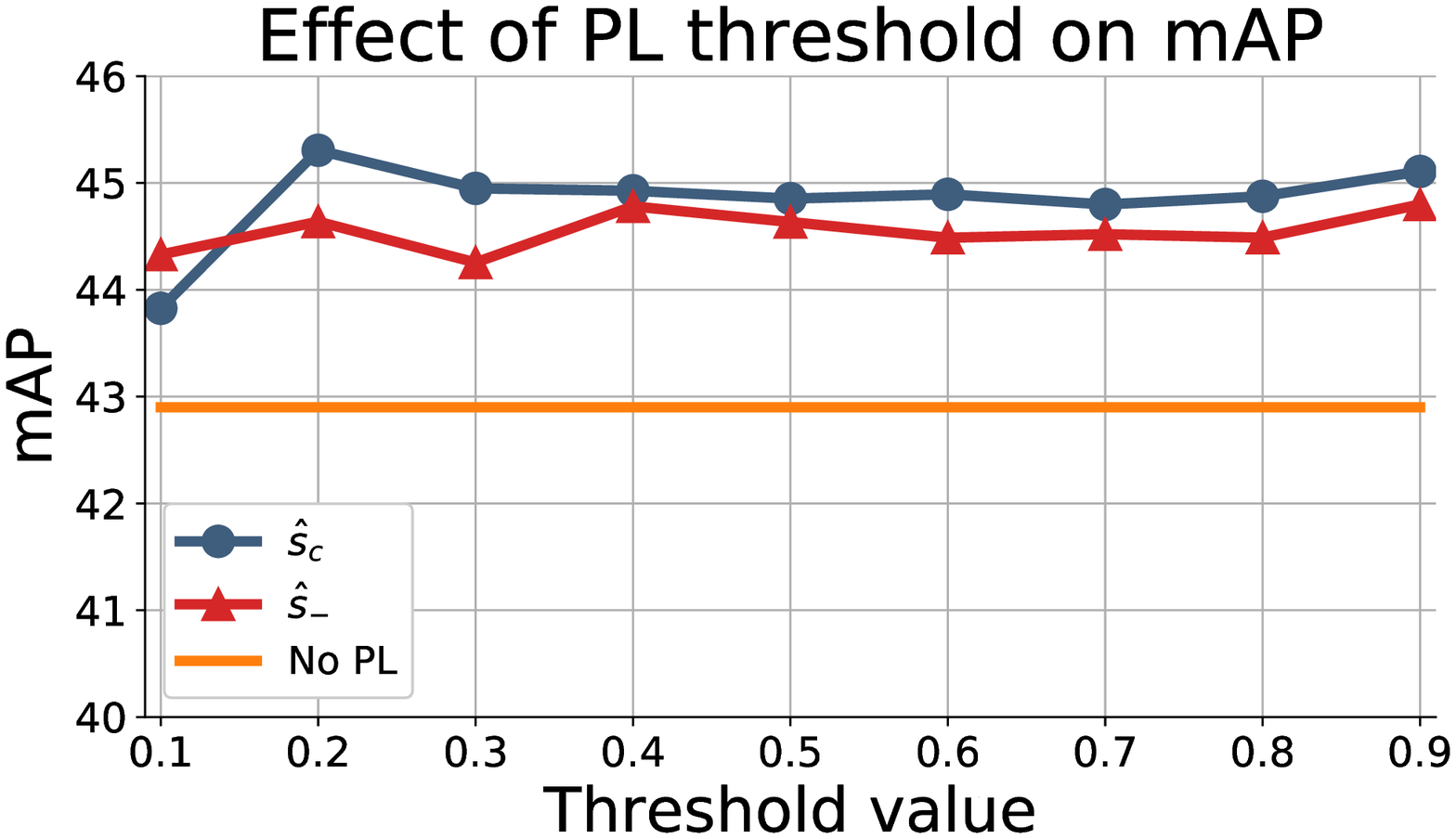}
	\end{subfigure}
	\begin{subfigure}[b]{0.49\linewidth}\centering
	\includegraphics[width=0.9\linewidth]{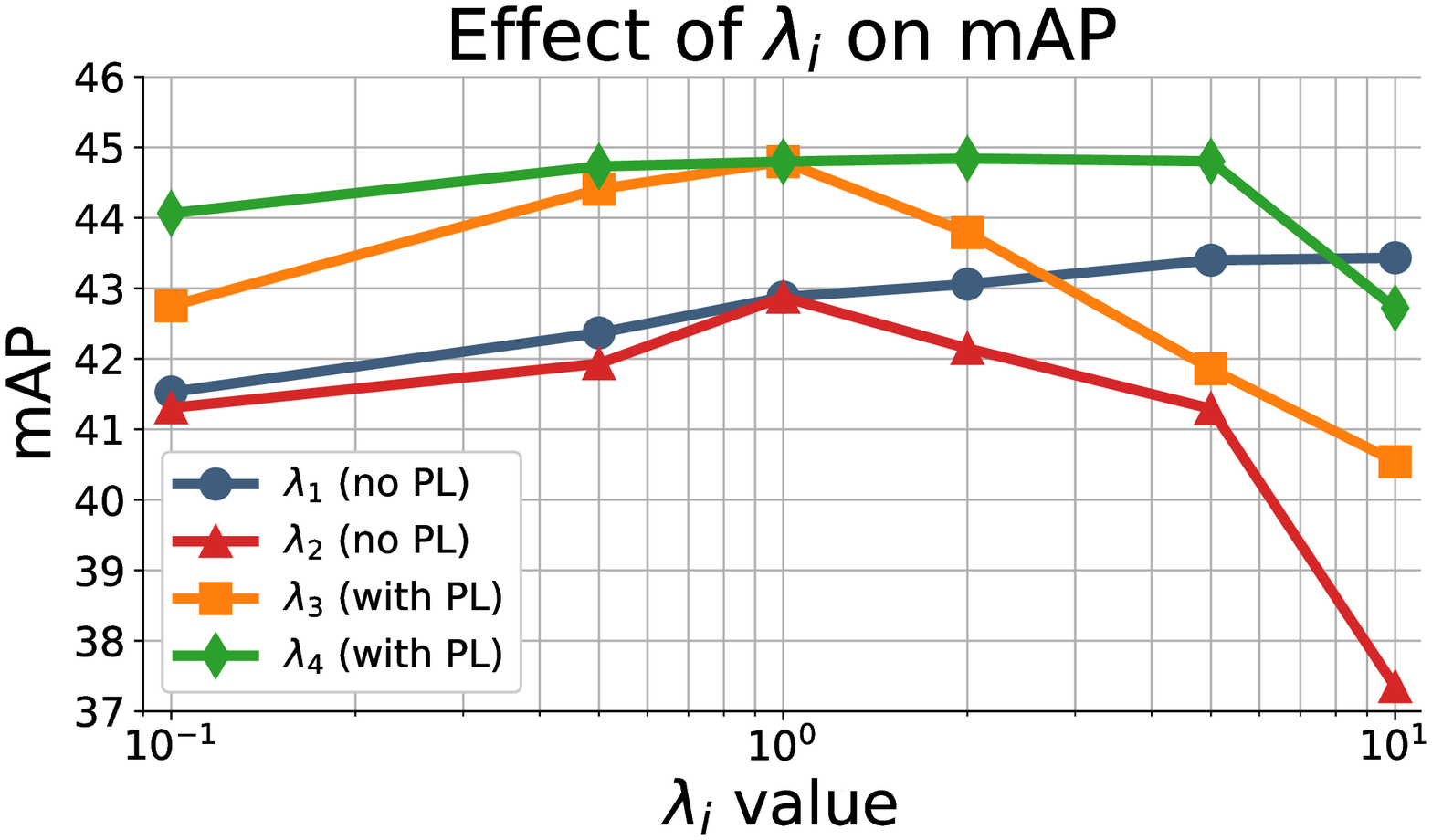}
	\end{subfigure}
	\vspace{-0.2cm}
	\caption{Left: Effect of the threshold used for pseudo labelling in Clipart1k. Right: Effect of varying $\lambda_i$ values in Clipart1k.}
	\label{fig:hyperparam_th}
	\vspace{-0.21cm}
\end{figure}

\noindent\textbf{Hyperparameter sensitivity.} Figure \ref{fig:hyperparam_th} shows the effect of varying $\hat{s}_c$ and $\hat{s}_-$ thresholds for pseudo labelling (see Section \ref{sec:pl}) in Clipart1k. The performance is stable for $\hat{s}_c\ge 0.2$ and low values of $\hat{s}_-$ give slightly lower performance. Additionally, for the standard hard-negative mining case, \ie $\hat{s}_-=0$, we obtained 44.0\%, that is 0.8\% lower compared to ${\hat{s}_-=0.9}$, showing the importance of carefully selecting the negative samples.  Figure \ref{fig:hyperparam_th}  also shows the effect of varying $\lambda_i$ in Clipart1k (see Section \ref{sec:combined}). For the first step of training, we vary the value of $\lambda_1$ and $\lambda_2$, which are the weights for $\mathcal{L}_{\mathit{ST}}$ and $\mathcal{L}_{Cons}$. We find that the performance increases when the style loss is given more weight. As for $\mathcal{L}_{Cons}$,  $\lambda_2=1.0$ achieves the best result and $\lambda_2=10.0$ is the only value that drops the performance compared to $\lambda_2=0.0$ (see Table \ref{tab:results_clipart}). For the second step, we vary the weight of $\mathcal{L}_{\mathcal{S}}$ and $\mathcal{L}_{\mathit{RPL}}$ in the loss finding that the optimal setting is weighting them equally with $\lambda_3=1.0$ and $\lambda_4=1.0$.\par

\noindent\textbf{Driving datasets.} We follow the same protocol in \cite{saito2017asymmetric} to test our method in two experiments: adaptation from Sim10k \cite{johnson2016driving} (simulated data) to Cityscapes~\cite{cordts2016cityscapes} (real data) and from Cityscapes to Foggy Cityscapes \cite{Sakaridis18foggy} (synthetic fog added to Cityscapes). A detailed overview of the datasets is included in the supplementary material. We employed the SSD512 model, which uses images of 512x512 pixels, due to small objects being present in the images. The batch size is reduced to 2 due to memory constraints, and for Cityscapes$\rightarrow{\text{Foggy Cityscapes}}$ we doubled the number of iterations (20000 and 10000 iterations for first and second step). Table \ref{tab:results_car} shows the results. We also report the results given in \cite{saito2018strong} when using only source data for the Faster R-CNN method due to the difference with SSD512. The improvement is limited for Sim10k$\rightarrow{\text{Cityscapes}}$, whereas in Cityscapes$\rightarrow{\text{Foggy Cityscapes}}$ we see an improvement of 4.1\% over the source domain trained model. Style transfer adaptation is greatly beneficial for Cityscapes$\rightarrow{\text{Foggy Cityscapes}}$ as Foggy Cityscapes only differs from Cityscapes in the low-level statistics, \ie added fog. Consistency only helps when using $\mathcal{T}$ instances, where it increases the performance by 0.7\% compared to not applying it. In these datasets, the improvements are limited for several reasons. First, the target datasets do not have a large variation in style in the instances, which also limits the benefits of forcing feature consistency among the different styles. Second, the low-level feature mismatch is lower compared to Pascal VOC and the cross-domain dataset. Lastly, the style transfer method is not suited to generate realistic images and the output images contain distortions (examples of style-transferred images included in the supplementary material). Using methods adapted for realistic images such as \cite{li2018closed, yoo2019photorealistic} may improve the performance. In both experiments, the best result is obtained when all of the proposed blocks are used. Notably, using pseudo labels without $x_{\mathcal{T}\rightarrow{\mathcal{T}}}$ instances makes the performance drop in both experiments, \red{showing that forcing consistency in the detections between $x_{\mathcal{T}}$ and $x_{\mathcal{T}\rightarrow{\mathcal{T}}}$ instances is also beneficial in this setting}.

\section{Conclusion}
We presented an approach for domain adaptation of an object detector based on low-level adaptation via style transfer and high-level adaptation via robust pseudo labelling. We demonstrated that each of these contributions brings an improvement with the highest performance obtained when all are combined. Our approach gives the highest mean Average Precision in the cross-domain dataset, where we proved the benefits of generating several translated source images to any of the target styles. We also showed that using style transfer with pseudo labelled instances is also key to improve the results. The method was also tested in driving datasets using the same parameters, showing an improvement over the source trained model.

\noindent{\bf Acknowledgment.} This research was supported by
FACER2VM EPSRC EP/N007743/1.

\bibliography{egbib}
\end{document}